\documentclass{article}

\usepackage[preprint]{neurips_2022}

\usepackage[utf8]{inputenc} 
\usepackage[T1]{fontenc}    
\usepackage{hyperref}       
\usepackage{url}            
\usepackage{booktabs}       
\usepackage{amsfonts}       
\usepackage{nicefrac}       
\usepackage{microtype}      
\usepackage{xcolor}         

\usepackage{times}
\usepackage{latexsym}

\usepackage[T1]{fontenc}

\usepackage[utf8]{inputenc}

\usepackage{microtype}

\usepackage{inconsolata}

\usepackage{times}
\usepackage{latexsym}
\usepackage{graphicx}
\usepackage{multirow}
\usepackage{tikz}
\usepackage{amsmath}
\usepackage{diagbox}
\usepackage{amssymb}
\usepackage{makecell}
\usepackage{bm}
\usepackage{hyperref}
\usepackage{soul}

\usepackage{enumerate}
\usepackage{algorithmic}
\usepackage{subfigure}
\usepackage{graphicx}
\usepackage{amsmath}
\usepackage{subfigure}

\usepackage{amsfonts,amssymb}
\usepackage{nicefrac}       
\usepackage{multicol}
\usepackage{multirow}
\usepackage{color}

\usepackage{pgfplots}
\usepackage{pgfplotstable}
\pgfplotsset{width=7cm,compat=1.13}
\usepackage[T1]{fontenc}
\usepackage{bbding}
\usepackage{balance}

\usepackage{algorithm}
\usepackage{color}

\title{\textsc{FewFedWeight}: Few-shot Federated Learning Framework across Multiple NLP Tasks}

\author{
Weilong Dong\textsuperscript{1}\footnotemark[2] \footnotemark[3], Xinwei Wu\textsuperscript{1}\footnotemark[2] \footnotemark[3], Junzhuo Li\textsuperscript{2}\footnotemark[3], \\
\bf{Shuangzhi Wu\textsuperscript{3}, Chao Bian\textsuperscript{3}, Deyi Xiong\textsuperscript{1}\textsuperscript{2} \footnotemark[1]}\\
        \textsuperscript{1}College of Intelligence and Computing, Tianjin University, Tianjin, China \\
        \textsuperscript{2}School of New Media and Communication, Tianjin University, Tianjin, China \\
        \textsuperscript{3}ByteDance Lark AI, Beijing, China \\
        \texttt{\{willowd, wuxw2021, jzli, dyxiong\}@tju.edu.cn,} \\ 
        \texttt{wufurui@bytedance.com, chaobian@outlook.com
        }
}

\begin{document}

\maketitle

\begin{abstract}
  Massively multi-task learning with large language models has recently made substantial progress on few-shot generalization. 
However, this is usually performed in a centralized learning fashion, ignoring the privacy sensitivity issue of (annotated) data used in multiple tasks.
To mitigate this issue, we propose \textsc{FewFedWeight}, a few-shot federated learning framework across multiple tasks, to achieve the best of both worlds: privacy preservation and cross-task generalization. 
\textsc{FewFedWeight} trains client models in isolated devices without sharing data. 
It broadcasts the global model in the server to each client and produces pseudo data for clients so that knowledge from the global model can be explored to enhance few-shot learning of each client model. 
An energy-based algorithm is further proposed to weight pseudo samples in order to reduce the negative impact of noise from the generated pseudo data.
 Adaptive model weights of client models are also tuned according to their performance.
We use these model weights to dynamically aggregate client models to update the global model. 
Experiments on 118 NLP tasks show that \textsc{FewFedWeight} can significantly improve the performance of client models on 61\% tasks with an average performance improvement rate of 30.5\% over the baseline and substantially outperform FedAvg \citep{mcmahan2017communication} and other decentralized learning methods.
\end{abstract}

\renewcommand{\thefootnote}{\fnsymbol{footnote}}
\footnotetext[1]{Corresponding author.}
\footnotetext[2]{Equal contribution.}
\footnotetext[3]{Work done while this author was an intern at BtyeDance Lark AI.}

\section{Introduction}

Large language models \citep{brown2020language,liu2021pre} (LLMs) are capable of performing few-shot learning on downstream tasks, where a small number of annotated instances are provided to teach the model a new task \citep{liu2021pre,liu2021p,du2022glm}.
Significant interest has recently emerged on transforming a wide array of downstream tasks into a unified form, e.g., human-readable prompts or instructions \cite{sanh2021multitask, wei2021finetuned}, and continuing to train/finetune LLMs on the transformed data so as to endow LLMs with few/zero-shot task generalization \citep{sun2021nsp,ye2021crossfit,xu2022zeroprompt}.
The goal of such efforts might be towards general linguistic intelligence that can reuse previously acquired linguistic knowledge and adapt to a new task quickly \citep{yogatama2019learning}.
Intuitively and empirically, cross-task generalization benefits from massive and diversified downstream tasks \cite{sanh2021multitask, wei2021finetuned}.

However, in real-world scenarios, the data of different tasks are usually distributed across isolated clients (e.g., devices, users or institutions) and privacy-sensitive.
Local private data are not legally allowed to leave their owners/clients by law, e.g., EU GDPR \cite{voigt2017eu}. 
Owing to privacy risk or legality, collecting sufficient multi-task data for centralized training becomes difficult. 
To address this issue, Federated Learning (FL) \citep{konevcny2016federated, mcmahan2017communication} is proposed, where each client participates in a training process without sharing private data with other clients.

\begin{figure*}[t]
    \includegraphics[width=0.95\textwidth]{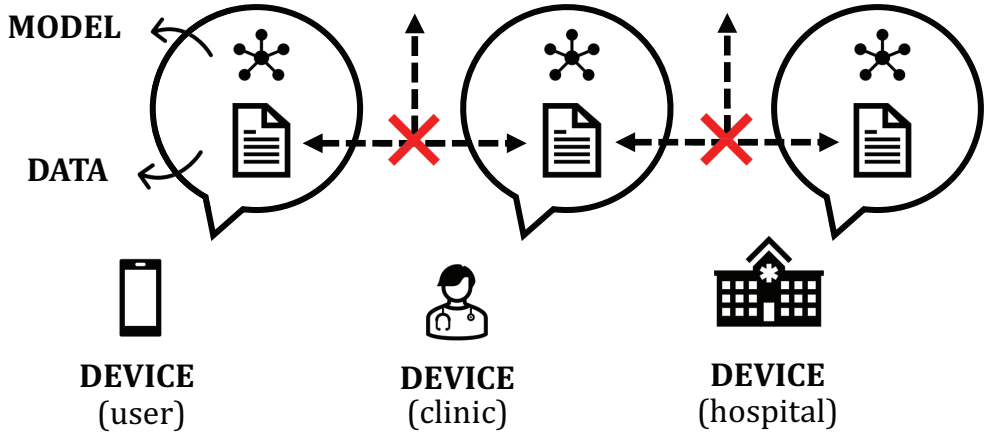}
    \caption{Isolated multi-task data in local devices.}
    \label{fig:fedavg}
\end{figure*}

In this paper, our key interest is to combine federated learning with multi-task learning in the few-shot setting, where each client owns a small amount of private data (illustrated in Figure \ref{fig:fedavg}), in the goal of achieving the best of both worlds: cross-task generalization and data privacy. 
This scenario is pervasive, for example, different hospitals own a small number of annotated privacy-sensitive examples on different tasks (e.g., medical QA, named entity recognition or relation extraction over electronic health records (EHR), EHR summarization, medical image captioning).  
The challenges of this few-shot federated learning across multiple tasks are at least two-fold: (1) teaching models to few-shot learn and (2) sharing cross-task knowledge across clients without sharing private data. 

To deal with the aforementioned research questions, we propose a weighted few-shot federated learning framework (\textsc{FewFedWeight}) for massive tasks to harness LLM-based few-shot learning with federated learning. 
Specifically, we construct a global model at the server, which aggregates information from client models located in each client and facilitates knowledge transfer across clients and tasks. 
In order to few-shot learn efficiently in each client, we use the global model to produce pseudo labeled data for each client model. 
To reduce the negative impact from noise in the generated pseudo data, we further propose an energy-based algorithm to weight the generated examples. 
In order to aggregate the trained client models into the global model, we introduce a dynamic aggregation method, which estimates the weights of client models according to the performance of the corresponding client model on the pseudo examples generated by the global model and the annotated examples.

The main contributions of our work are summarized as follows:
\begin{itemize}
    \item We propose a federated learning framework for multi-task learning in the few-shot and private setting, which is equipped with two components: an energy-based weighting algorithm for updating the weights of pseudo examples generated by the global model and a dynamic aggregation method based on the performance of client models. 
    \item We conduct experiments on 118 open-source tasks to evaluate the proposed \textsc{FewFedWeight} framework. To the best of our knowledge, this is the first attempt to perform federated learning on massive NLP tasks.
    \item Experiments demonstrate that (1) \textsc{FewFedWeight} significantly outperforms the classic federated learning FedAVG, the personalized federated learning Ditto \cite{li2021ditto}, centralized training, data silo data augmentation and meta weighting based data augmentation and (2)  the client models trained by \textsc{FewFedWeight} gain 30.5\% over the original model and they can generalize well cross multiple tasks.
\end{itemize}

\section{Preliminary: Federated Learning}

\begin{figure*}[t]
    \includegraphics[width=\textwidth]{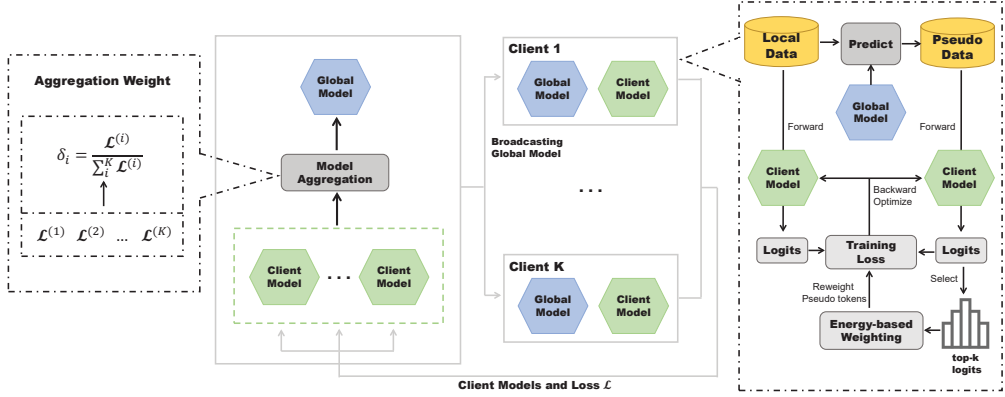}
    \caption{The diagram of \textsc{FewFedWeight}.}
    \label{fig:overview}
\end{figure*}

In federated learning, each client has its own private data which are not shared during training. Let $\mathbb{U} = \{u_1, u_2, ..., u_K\}$ be a set of clients, where $K$ is the number of clients. Apart from the clients, there is a central server. The server initializes a global model, sends the global model to a subset of clients for updating and aggregates the updated client models into the global model. We denote the server as $\mathcal{S}$. In the classic FL framework FedAvg \citep{mcmahan2017communication}, $\mathcal{S}$ aggregates the client models as follows:
\begin{equation} \label{eq:aggregation}
    \bm{F}_{\text{global}} = \sum_{i=1}^N \bm{\delta}_i\cdot \bm{F}_i,
\end{equation}
where $\bm{F}_\text{global}$ is the aggregated global model, $\bm{F}_i$ is the local model on client $u_i$ and $\bm{\delta}_i$ is the model weight for $\bm{F}_i$. In FedAvg, the model weight $\bm{\delta}$ is estimated according to the amount of data on each client, which is computed as $\bm{\delta}_j = \frac{n_j}{\sum_{j=1}^{K} n_j}$, where $n_j$ is the number of training samples on $u_j$.

In addition to the global federated learning methods, prior works \citep{smith2017federated, marfoq2021federated} have explored personalized federated learning (PFL) for multi-task learning. In PFL, all clients jointly train a global model, but each client can built its own personalized client model by interpolating the global model and its client model. During training, a personalized client model is able to exploit information of other clients from the global model without sharing their data. A common method in PFL is to add a regularization term in the loss function. Specifically, the loss of each client model is computed as $\mathcal{L}_i = \hat{\mathcal{L}}_i + ||\bm{F}_\text{global}-\bm{F}_i||^2$, where $\hat{\mathcal{L}}_i$ is the original loss on $u_i$. We choose this PFL setting in our work for tackling different tasks.

\section{Methodology}
\label{methods}

In this section, we introduce our proposed \textsc{FewFedWeight} for few-shot federated learning across multiple tasks. The architecture of our module is illustrated in Figure \ref{fig:overview}. We first overview \textsc{FewFedWeight} (\S \ref{overview}) and then elaborate the two essential components in our framework: data augmentation with energy-based sample weighting (\S \ref{generator}) and dynamic model aggregation (\S \ref{model_reweight}).

\subsection{Overview} \label{overview}

Algorithm \ref{alg:overview} formalizes the entire flow of \textsc{FewFedWeight}. We denote the training data as $\{(\bm{X}, \bm{Y})\}, \bm{X}=\{\bm{X}_1,...,\bm{X}_K\}, \bm{Y}=\{\bm{Y}_1,...,\bm{Y}_K\}$, where $(\bm{X}_i$, $\bm{Y}_i)$ is the training data on client $u_i$. 

At the beginning, the server $\mathcal{S}$ uses a pre-trained language model as the initial global model $\bm{F}^{(0)}_{\text{global}}$ and defines the number of training epochs $T$. $\mathcal{S}$ broadcasts $\bm{F}^{(0)}_{\text{global}}$ to each client and the initial client model $\bm{F}_i^{(0)}$ on $u_i$ is initialized as $\bm{F}^{(0)}_{\text{global}}$. 

In the $t$-th training epoch, $\mathcal{S}$ sends $\bm{F}^{(t)}_{\text{global}}$ to each client. Instead of replacing $\bm{F}^{(t)}_{i}$ with $\bm{F}^{(t)}_{\text{global}}$, $\bm{F}^{(t)}_{\text{global}}$ is used to predict pseudo labels $\hat{\bm{Y}}_i$ on $\bm{X}_i$, which will be described in subsection \ref{generator} (Line 7 in Algorithm \ref{alg:overview}). 
Then an energy-based weighting algorithm is performed to evaluate $\hat{\bm{Y}}_i$'s quality and estimate weights $\bm{w}_\text{pseudo}$. With the estimated weights, the client model $\bm{F}_i^{(t)}$ is updated on the combination of $(\bm{X}_i, \bm{Y}_i)$ and $(\bm{X}_i, \hat{\bm{Y}}_i)$ (Line 8 in Algorithm \ref{alg:overview}). 
The updated model $\bm{F}_i^{(t+1)}$ and training loss are sent to $\mathcal{S}$ for model aggregation. The server $\mathcal{S}$ computes the new global model $\bm{F}^{(t+1)}_{\text{global}}$, which will be elaborated in subsection \ref{model_reweight} (Line 5 in Algorithm \ref{alg:overview}).

\begin{small}
\begin{algorithm}[t]
	\caption{\textsc{FewFedWeight}}
	\label{alg:overview}
    \textbf{Input}: $T, \{(\bm{X}, \bm{Y})\}$, $\bm{F}^{(0)}_{\text{global}}$ \\
	\begin{algorithmic}[1] 
		\FOR{$t = 0,1,2,...,T-1$}
		\STATE \textbf{Server:}
		\STATE Send $\bm{F}^{(t)}_{\text{global}}$ to clients
		\STATE $\bm{F}_i^{(t+1)}, \mathcal{L}^{(i)} \gets\text{Client}(\bm{F}^{(t)}_{\text{global}}, \bm{X}_i, \bm{Y}_i)$
		\STATE Compute $\bm{F}^{(t+1)}_{\text{global}}$ based on Eq. (\ref{eq:aggregation}) \& (\ref{eq:model_weight})   
		\STATE \textbf{Client:}
        \STATE Generate pseudo labels 
        \item[] \qquad $\hat{\bm{Y_i}}=\text{Predict}(\bm{F}_i^{(t)}, \bm{X}_i)$
		\STATE  $\bm{F}_i^{(t+1)}, \mathcal{L}^{(i)} \gets $    
		\item[] \qquad $\text{Client Model Updating}(\bm{F}_i^{(t)},\bm{X}_i,\bm{Y}_i,\hat{\bm{Y}}_i)$
		\STATE \textbf{return} $\bm{F}_i^{(t+1)}, \mathcal{L}^{(i)}$
		\ENDFOR
	\end{algorithmic}
\end{algorithm}
\end{small}

\subsection{Data Augmentation} \label{generator}

\paragraph*{Pseudo Data Generation with Global Knowledge} Data augmentation can ease the data scarcity issue in few-shot learning \citep{feng2021survey}. In our case, each client can synthesize pseudo data using its own client model. However, this cannot allow each client model to leverage information of other clients and the client model trained on its own data is not able to generalize across other tasks. Hence, we use the global model to synthesize data for each client.

Suppose the unlabeled data on $u_i$ is $\bm{X}_i=\{\bm{x}_1^{(i)}, \bm{x}_2^{(i)}, ... , \bm{x}_{n_i}^{(i)}\}$ and corresponding label is $\bm{Y}_i =\{\bm{y}_1^{(i)}, \bm{y}_2^{(i)}, ... ,\bm{y}_{n_i}^{(i)}\}$. Following \citet{raffel2020exploringa} and \citet{ye2021crossfit}, where different tasks are unified into the same text-to-text format to enable knowledge transfer across tasks, we define a prompt template for task conversion and unification: \textit{<task name>: <$\bm{x}_i$> [SEP] <$\bm{y}_i$>}. Take question answering as an example, $\bm{x}_i$ is reformulated as: \textit{<question> [SEP] <choices>}, $\bm{y}_i$ is the corresponding answer.

At the $t$-th FL training epoch, $u_i$ receives the global model $\bm{F}_{\text{global}}^{(t)}$ from the server $\mathcal{S}$ and uses $\bm{F}_{\text{global}}^{(t)}$ to predict the labels of $\bm{X}_i$. We denote the predicted pseudo labels as $\hat{\bm{Y}}_i =\{\hat{\bm{y}}_1^{(i)}, \hat{\bm{y}}_2^{(i)}, ... ,\hat{\bm{y}}_{n_i}^{(i)}\}$. When predicting $\hat{\bm{Y}}_i$, $\bm{F}_{\text{global}}^{(t)}$ actually performs the task with $\bm{X}_i$ being fed as input. The generated output with the highest probability will be taken as $\hat{\bm{Y}}_i$.

The predicted pseudo data are then combined with the original local data to train the client model. On the training instances from the pseudo data $\{\bm{X}_i, \hat{\bm{Y}}_i\}$, the client model is updated with an energy-based weighting algorithm to minimize a loss $\mathcal{L}^{(i)}_\text{pseudo}$, while on the instances from the original local data $\{\bm{X}_i, \bm{Y}_i\}$, the client model is trained as usual to optimize a loss $\mathcal{L}^{(i)}_\text{annotated}$. 

\paragraph*{Energy-based Pseudo Instance Weighting} \label{sample_reweight}
Although the pseudo instances synthesized by the global model contain global knowledge, they are noisy. We hence propose an energy-based token-wise weighting method to reduce the negative impact of noise from these pseudo instances.

\citet{liu2020Energybased} introduce the following energy function to perform out-of-distribution detection in image classification:
\begin{equation} \label{energy-function1}
    E(\bm{x};\bm{f}) = -T\cdot \mathrm{log}\sum_{i}^{K}e^{\bm{f}_i(\bm{x})/T}
\end{equation}
where $T$ is the temperature parameter, $K$ is the number of image classes, $\bm{x}$ is the input image and $\bm{f}_i(\bm{x})$ is the output logit in dimension $i$ (i.e., the logit value for class $i$). The energy $E(\bm{x};\bm{f})$ is higher for unobserved samples and lower for observed ones. 

\begin{figure}[t]
    \centering
    \includegraphics[width=0.8\textwidth]{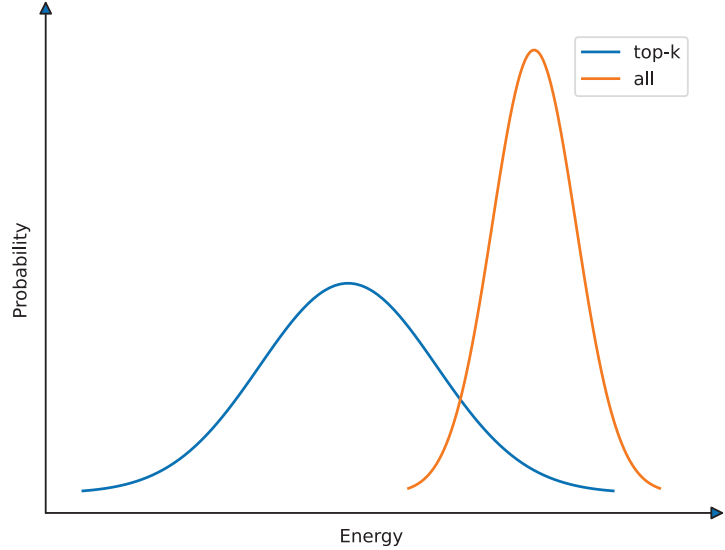}
    \caption{Energy distributions over top-k logits and all logits. The energy distribution over all logits (orange curve) is "crowded" and the energy values for different tokens are almost indistinguishable from each other.}
    \label{fig:energy}
\end{figure}

The essence of the energy function indicates that noisy instances have higher energy than clean instances. 
However, adapting Eq. (\ref{energy-function1}) to our model is nontrivial as we are confronted with two challenges. First, our task is a generation task rather than classification, which means that the number of sequences (classes) grows exponentially with sequence length. In order to deal with this, we estimate token-wise energy instead of sequence-wise energy as token-wise generation can be treated as a multi-class classification problem (prediting a word from the vocabulary). Second, due to the large size of the vocabulary used in our model, 
energy for different tokens calculated by Eq. (\ref{energy-function1}) would be almost indistinguishable from each other, as shown in Figure \ref{fig:energy}.

Although the dimension of logits is large, only logits with high values have a substantial impact on the final prediction. We hence choose the top-k logits to calculate the energy as follows:
\begin{equation} \label{energy-function2}
    E(\bm{a};\bm{f}) = -T\cdot \mathrm{log}\sum_{i}e^{\bm{f}_i(\bm{a})/T}, i\in \mathbb{K}
\end{equation}
where $\bm{a}$ is the predicted token, $\bm{f}(\bm{a})$ denotes the logits corresponding to $\bm{a}$, $\mathbb{K}$ is the set of indices of logits with the top-k values. 
A noisy token will have larger energy and should have a lower weight. The instance weight is hence computed by normalizing token-wise energy at the sequence level:
\begin{equation} \label{token-weight}
    \bm{w}_{\text{pseudo}}^{\bm{a}} = \frac{-E(\bm{a};\bm{f})}{\sum_i^l -E(\bm{i};\bm{f})}
\end{equation}
where $l$ is the length of a predicted pseudo label. The training loss for this pseudo label is hence defined as:
$\mathcal{L}_{\hat{y}}=\sum_i^l \bm{w}_{\text{pseudo}}^i\cdot \mathcal{L}_{\hat{y}}^i$, where $\mathcal{L}_{\hat{y}}^i$ is the NLL loss of token $i$.

\subsection{Dynamic Aggregation} \label{model_reweight}

Aggregation weights in FL 
\citep{mcmahan2017communication,li2020federated} and PFL \citep{smith2017federated, li2021ditto} are usually estimated according to the number of training instances in clients, which is not adaptive to the training performance of client models. Since different clients have different tasks and the degree of difficulty to learn is varying across tasks, it would be beneficial if aggregation weights could match the task difficulty of clients (i.e., paying more attention to harder tasks).  

To achieve this purpose, we estimate the aggregation weight $\bm{\delta_i}$ in Eq. (\ref{eq:aggregation}) as follows:
\begin{equation} \label{eq:model_weight}
    \bm{\delta}_i = \frac{\mathcal{L}^{(i)}}{\sum_i^K \mathcal{L}^{(i)}}, \\
\end{equation}
where $\mathcal{L}^{(i)}=\mathcal{L}^{(i)}_{\text{pseudo}}+\mathcal{L}^{(i)}_{\text{annotated}}$, which is the total training loss on client $u_i$, the sum of the loss on the weighted pseudo samples and that on the annotated samples.

The weight $\bm{\delta_i}$ for client model $\bm{F}_i$ is proportional to $\bm{F}_i$'s training loss. 
A large weight suggests that tasks in client $u_i$ are challenging. 
Similar to the essence of AdaBoost \citep{freund1996experiments}, 
the client model with a larger weight (i.e., with poorer performance) plays a more important role in model aggregation and thus the challenging tasks are paid more attention to.

\section{Experiments}
We examined the effectiveness of the proposed \textsc{FewFedWeight} on a wide range of tasks. 

\subsection{Setup} \label{exp:details}

\paragraph*{Datasets} 
We followed \citet{ye2021crossfit} to use 118 tasks from the Huggingface Datasets \citep{lhoest2021datasets}, which are categorized into 4 groups: classification, QA, conditional generation and other.
All tasks were reframed into the same text-to-text format.
The number of clients was set to 4 by default.
The numbers of tasks in the clients were set as evenly as possible: 29, 29, 29, 31, respectively. 
Each client had randomly selected and diversified tasks so that we could evaluate the cross-task generalization performance in the client models.
Details of the selected tasks and task distribution over clients are shown in Appendix \ref{appendix:split}. 

\paragraph*{Few-shot Setting} We used the few-shot sampling method of \citet{gao2021making} for our few-shot learning. 
For generation tasks (e.g., summarization, dialogue), 32 training examples were selected for each task. 
For classification tasks  (e.g., natural language inference, sentiment analysis), 16 examples were selected for each task. 
Since labeled data are rare in real-world scenario, our test sets also followed the same  few-shot setting as the training sets. 
To eliminate random factors, we chose 5 different random seeds to select few-shot samples. These selected samples collectively construct the training set and test set.

\paragraph*{Training Details} 
We chose BART-Base \citep{lewis2020bart} (140M parameters) as our client and global model. 
The training hyperparameters for \textsc{FewFedWeight} were set as follows: 20 as the number of training epochs, 3e-5 for learning rate with a linear warm up, 8 for the batch size. The top-7 largest logit values were selected to compute the energy and the temperature parameter for the energy function was set to 1.

\paragraph*{Baselines}
We compared the proposed \textsc{FewFedWeight} against the following baselines:
\begin{itemize}
    \item \textbf{Data Silo:} Each client trains a model with its own private data. This method serves as the base model for PIR evaluation. 
    \item \textbf{Data Silo DA:} Data augmentation is used for the Data Silo training and the numbers of training samples after data augmentation in clients are the same as those in \textsc{FewFedWeight}. 
    \item \textbf{Centralized Training:} We train a central model with all data put together. Note that the Centralized Training does not take data privacy into account, which is to show the oracle result at the cost of data privacy. 
    \item \textbf{FedAvg} \citep{mcmahan2017communication}\textbf{:} A classical FL algorithm, in which clients participate in training a global model but there are no personalized client models.
    \item \textbf{Ditto} \citep{li2021ditto}\textbf{:} A general framework for personalized federate learning, which achieves personalization by regularizing client models towards their average. Ditto mainly focuses on fairness and robustness in FL instead of the limited data and task heterogeneity.
    \item \textbf{Meta Weighting}\textbf{:} We follow \citet{ren2018learning} to use a meta-learning based method to weight pseudo samples in our data augmentation component (see Section \ref{generator}) for each client. This meta weighting approach serves as a strong baseline to compare with our energy-based weighting. 
    More details of the meta-learning based weighting are provided in Appendix \ref{appendix:metaweight}.
\end{itemize}

\paragraph*{Evaluation Metrics} Evaluation metrics for different tasks vary widely. Averaging metrics of different tasks hence doesn't make sense. 
In order to evaluate the performance of \textsc{FewFedWeight} in comparison to baselines and tackle the problem of incompatable metrics, we define a unified metric: average performance improvement rate (APIR).
The performance improvement rate (PIR) refers to how much the performance has been improved over the base model. The PIR is computed as follows:
\begin{equation}
    \text{PIR} = \frac{m_{\text{new}}-m_{\text{base}}}{m_{\text{base}}},
\end{equation}
where $m_{\text{new}}$ refers to the performance of the new model 
while $m_{\text{base}}$ is the performance of the base model in terms of the widely-used metric for the corresponding task. 
In our experiments, the base model refers to Data Silo by default. 
The APIR is the average of PIRs across all tasks. 
Each client has its own APIR and the APIR averaged over the four clients is reported as the APIR for the corresponding method. 

\begin{table}[t]
\centering
\small
\resizebox{0.8\textwidth}{!}{
\begin{tabular}{lllll}
\hline
\textbf{Method} & \textbf{APIR} & \textbf{Win} & \textbf{Lose} & \textbf{Tie} \\
\hline
Data Silo DA  & 8.2\% & 44 & 29 & 45\\
FedAvg & 5.1\% & 40 & 41 & 37\\
Ditto & 14.1\% & 45 & 34 & 39\\
Meta Weighting & 22.6\% & 49 & 32 & 37 \\
\textsc{FewFedWeight} & \textbf{30.5\%} & \textbf{72} & \textbf{7} & \textbf{39} \\
\hline
Centralized Training \qquad\qquad\qquad\qquad\qquad & 36.9\% & 65 & 16 & 37 \\
\hline
\end{tabular}
}
\caption{APIR results of different methods. Win/Lose/Tie is the number of tasks on which the corresponding method is better/worse than or as good as the base model (i.e., data silo without data augmentation).}
\label{tab:main_exp}
\end{table}

\subsection{Overall Performance} \label{exp:efficiency}
The overall results are shown in Table \ref{tab:main_exp}. 
We observe that FedAvg is not able to well handle few-shot multi-task learning.
Although FedAvg leverages the data from all clients through federated learning, it performs just slightly better than Data Silo.
The AIPR of Data Silo DA demonstrates that data augmentation is beneficial to few-shot learning.
Although each client can only use its own private data in Data Silo DA, both its APIR and the number of Win tasks are better than those of FedAvg.
The recently proposed PFL framework, Ditto, achieves better APIR than FedAvg and Data Silo DA.
However, it is worse than Meta Weighting.
As a strong baseline, Meta Weighting achieves promising performance. Due to its global data augmentation and meta-learning based pseudo sample weighting, the dilemma between data scarcity in few-shot learning and noise in data augmentation can be alleviated to some extent. However, its time complexity is very high due to the meta-learning procedure shown in Appendix \ref{appendix:metaweight}. Additionally, it is worse than \textsc{FewFedWeight} in terms of APIR. 

\begin{figure*}[t]
    \centering
    \includegraphics[width=0.9\textwidth]{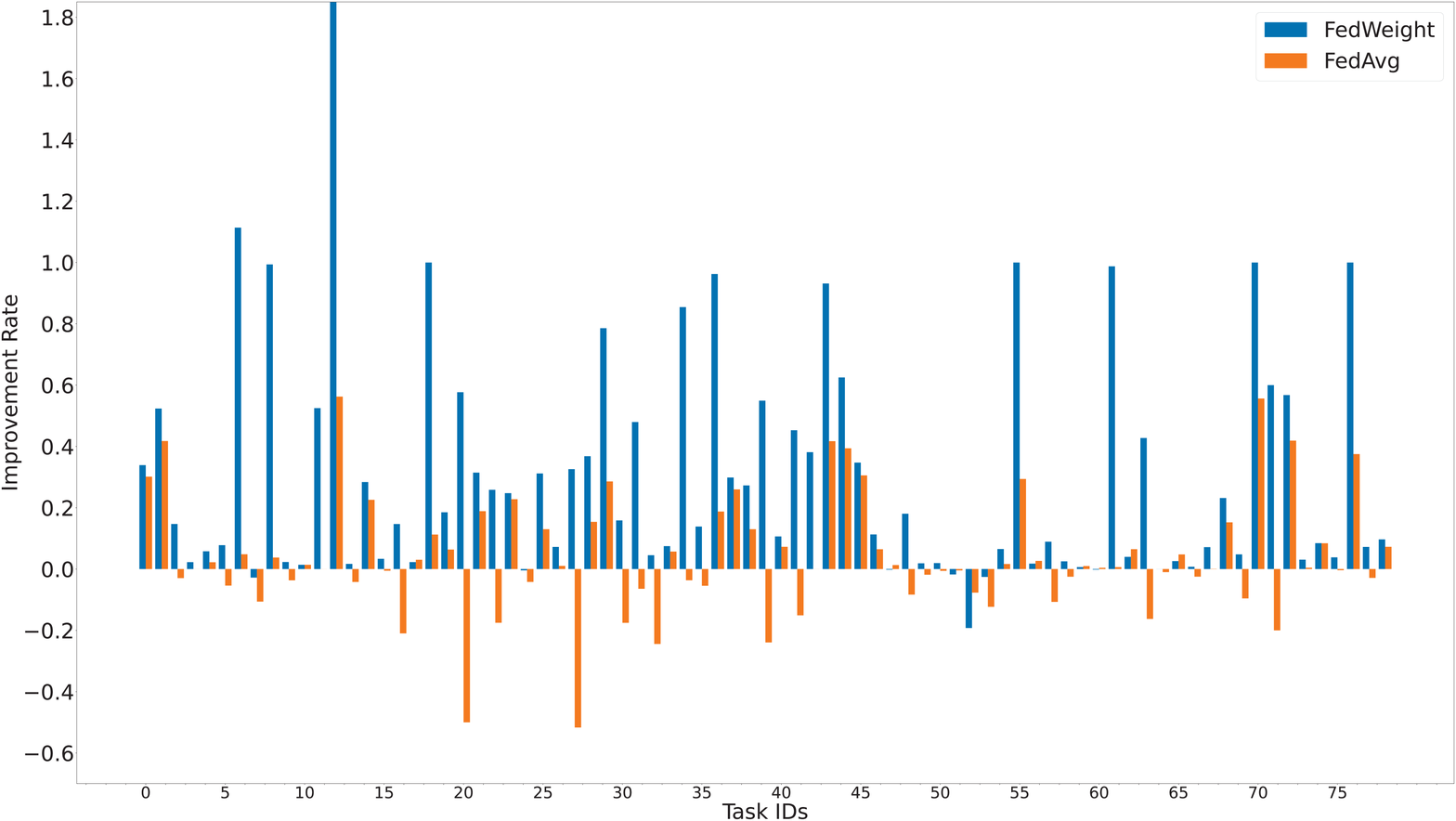}
    \caption{PIR results of FedAvg and \textsc{FewFedWeight} on different tasks.}
    \label{fig:tasks_improve_rate}
\end{figure*}

\textsc{FewFedWeight} achieves 30.5\% in APIR over Data Silo. 
Its APIR is close to that of Centralized Training and the number of Win tasks is even larger than that of Centralized Training. 
In order to have a clear comparison with classic FedAvg, we illustrate their PIRs on non-tie tasks in Figure \ref{fig:tasks_improve_rate}.

\begin{table}[t]
\centering
\resizebox{0.8\textwidth}{!}{
\begin{tabular}{lllll}
\hline
\textbf{Method} & \textbf{APIR} & \textbf{Win} & \textbf{Lose} & \textbf{Tie} \\
\hline
FedAvg + Data Augmentation \qquad\qquad\qquad & 11.5\% & 43 & 33 & 42\\
\quad + Pseudo Data Weighting & 27.0\% & 74 & 6 & 38\\
\quad + Dynamic Aggregation & 21.0\% & 64 & 16 & 38\\
\textsc{FewFedWeight} & \textbf{30.5\%} & \textbf{72} & \textbf{7} &\textbf{39}  \\
\hline
\end{tabular}}
\caption{Ablation study results of \textsc{FewFedWeight}.}
\vspace{-0.2cm}
\label{tab:ablation}
\end{table}

\subsection{Ablation Study} \label{exp:ablation}
\textsc{FewFedWeight} contains three essential components: data augmentation with global knowledge, 
energy-based pseudo data weighting and dynamic aggregation,
which collectively contribute to the advantage of \textsc{FewFedWeight}. We further conducted ablation experiments to investigate the effectiveness of the three components. Results are shown in Table \ref{tab:ablation}. 

FedAvg+Data Augmentation is similar to Data Silo DA in that both use augmented data, but the latter synthesizes pseudo data with its client model.
The APIR of FedAvg + DA is 3.3\% higher than that of Data Silo DA. We conjecture that this improvement could attribute to cross-task knowledge transfer brought by pseudo data synthesized by the global model.   

The energy-based pseudo data weighting significantly outperforms data augmentation by 15.5\% in APIR while the dynamic aggregation achieves an improvement of 9.5\% over data augmentation, suggesting that the two components are effective and beneficial to \textsc{FewFedWeight}. The three components together achieve further improvements over both the data augmentation + pseudo data weighting and data augmentation + dynamic aggregation.

\subsection{Varying the Number of Clients} \label{exp:clients_num}
To verify the stability of \textsc{FewFedWeight} over different numbers of clients, we conducted experiments in 4 settings with different numbers of clients. In order to have a fair comparison, we fixed the number of tasks on each client as the number of clients participating in FL training is varying. Five tasks were assigned to each client. We evaluated \textsc{FewFedWeight} on 2, 4, 8, 12 clients respectively. Results are reported in Table \ref{tab:clients_num}.

We observe that \textsc{FewFedWeight} has a remarkable stability across different numbers of clients. The APIR of 2 clients and 12 clients is slightly lower than that of the other two settings although they win in almost all tasks. 
We conjecture that increasing the number of clients can improve the number of win tasks as more information is incorporated. But it's hard to improve the APIR over all tasks as these tasks are distributed over more clients.

\begin{table}[t]
\centering
\tiny
\resizebox{0.8\textwidth}{!}{
\begin{tabular}{lllll}
\hline
\textbf{\#Clients       } & \textbf{APIR    } & \textbf{Win  
 } & \textbf{Lose   } & \textbf{Tie   } \\
\hline
2 \qquad\qquad & 23.2\% & 8 & 0 & 2\\
4 \qquad\qquad & 29.5\% & 12 & 5 & 3\\
8 \qquad\qquad & 29.8\% & 24 & 9 & 7\\
12 \qquad\qquad\qquad\qquad\qquad\qquad & 22.8\% & 37 & 9 & 14 \\
\hline
\end{tabular}
}
\caption{Results of different number of clients. The total number of tasks varies with the number of clients. }
\label{tab:clients_num}
\end{table}

\subsection{Training Cost} \label{exp:training_cost}
We further compared \textsc{FewFedWeight} with other baselines in terms of the training time and communication time between clients and the server. 
The additional training time of \textsc{FewFedWeight} mainly comes from the data augmentation component because it needs to predict pseudo labels for all samples in each epoch. The time for generating pseudo data is the same for all methods  with data augmentation (e.g., Data Silo DA, Meta Weighting). The training/generating time of one epoch over different methods is shown in Table \ref{tab:time_cost}.

\begin{table}[t]
\centering
\resizebox{0.8\textwidth}{!}{
\begin{tabular}{lllll}
\hline
\textbf{Methods} & \textbf{Generating} & \textbf{Training} & \textbf{Total} \\
\hline
FedAvg/Ditto  & 0 & 2 & 2\\
\textsc{FewFedWeight}/DataSilo DA & 13 & 4 &17\\
Meta Weighting & 13 & 8 & 21\\
\hline
\end{tabular}
}
\caption{The time cost of different methods on one epoch of the local training of a client (i.e., time units per epoch). 
}
\label{tab:time_cost}
\end{table}

\textsc{FewFedWeight} doesn't require extra time except for generating and training on pseudo data (The training time is twice as long as FedAvg because the size of total training data is doubled) while Meta Weighting requires more training time on pseudo data (details of meta-weighting is shown in Appendix \ref{appendix:metaweight}).

The communication cost of \textsc{FewFedWeight} is almost the same as FedAvg or Ditto, as the extra communication is brought by sending the training loss $\mathcal{L}^{(i)}$ to the server $\mathcal{S}$. Such transmission time is almost negligible compared to the training time of client model $\bm{F}_i$.

\subsection{Analysis on the Impact of Dynamic Aggregation} \label{exp:model_weight}

\begin{figure}[t]
    \centering
    \includegraphics[width=0.8\textwidth]{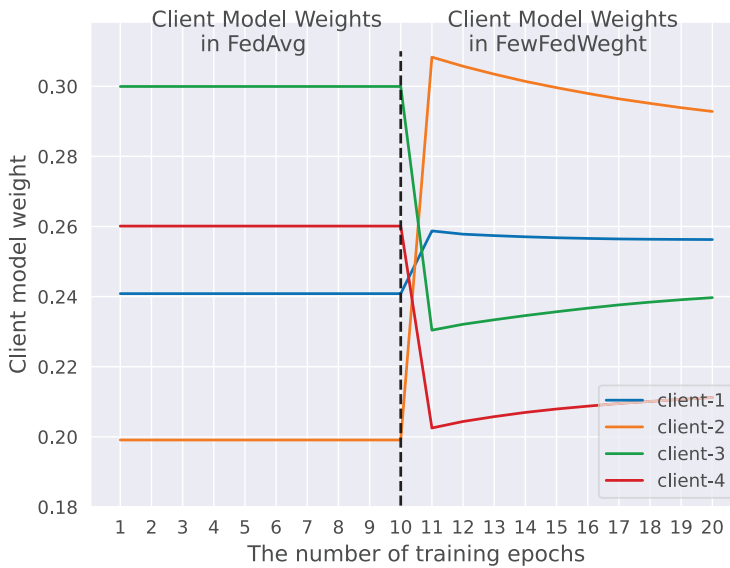}
    \caption{Client model weights learned by FedAvg vs. \textsc{FewFedWeight}.}
    \label{fig:model_weights}
\end{figure}

In order to look into the difference between FedAvg and \textsc{FewFedWeight}, we adopted the model weights of FedAvg in the first 10 training epochs while those of \textsc{FewFedWeight} are used in the last 10 training epochs. The curves of weights of the 4 client models during training are shown in Figure \ref{fig:model_weights}. 

In the first 10 epochs, FedAvg assigns the smallest model weight to client 2 because the amount of training data in client 2 is the smallest. 
When we use \textsc{FewFedWeight} in the second 10 epochs, the model weight of client 2 becomes the largest due to its worst performance.
Its weight then becomes lower, demonstrating that its performance is getting better gradually in the second 10 epochs. Additionally, the weights of the four client models tend to be stably changing as the number of training epochs increases.

\section{Related Work}

\paragraph*{Few-shot Multi-task Learning in NLP}
Large language models \citep{brown2020language,zhang2021cpm,perez2021true} achieve substantial progress on few-shot learning, thanks to the implicit multi-task learning in the pretraining of language models \cite{radford2019language, sanh2021multitask}.
A variety of efforts \citep{narayan2018don,raffel2020exploring,du2021all} have been devoted to mapping a wide range of NLP downstream tasks into a unified form. 
For example, \citet{narayan2018don} redefines the NLU tasks as generating specified answers through automatic prompts.
Natural language templates (prompts) \citep{liu2021p,du2022glm} have also been explored to reframe different NLP tasks into a unified text-to-text generation task.
Such efforts facilitate the standardization of few-shot multi-task learning process \citep{bragg2021flex, ye2021crossfit}.
Significantly different from these studies, we explore few-shot multi-task learning in a decentralized environment. 
\paragraph*{Data Augmentation}
Data augmentation is to automatically increase the amount and diversity of training data without explicitly collecting new data \citep{feng2021survey}. It has been widely used in computer vision and natural language processing to address issues such as low-resource language processing \citep{qin2021cosda}, bias mitigation \citep{zhao2018gender}, imbalanced data \citep{chawla2002smote}, few-shot learning \citep{lee2021neural}. In our work, we automatically generate pseudo data to alleviate the problem of insufficient data in local client training.

\paragraph*{Federated Learning}
Federated Learning (a.k.a collaborative learning) \citep{konevcny2016federated,chen2019federated} is a decentralized machine learning technique, where heterogeneous training data are stored locally on each client and are not exchangeable across clients during the training process.
Textual data are crucial for many NLP tasks, which may be privacy-sensitive, especially for those collected from edge devices (e.g., mobile phones). 
In this aspect, recent years have witnessed growing interest in the intersection of NLP and federated learning \citep{garcia2020decentralizing, liu2021federated}.

\paragraph*{FL with Few-shot or Multi-task Learning}
The inherent decentralization, heterogeneity and privacy-awareness in federated learning make it naturally suitable to unite FL with multi-task learning and few-shot learning \cite{zhang2021survey}. For few-shot federated learning, methods such as adversarial learning \citep{fan2021federated} or self-supervision \citep{shome2021fedaffect} have been exploited in the federated learning setting to learn from scarce data or use unlabeled data to alleviate the lack of supervised data.
And federated learning can be naturally embedded into the multi-task learning framework due to its distributed learning \citep{smith2017federated,marfoq2021federated,huang2021personalized}. 
In this aspect, \citet{smith2017federated} pioneer and frame federated multi-task learning with {\em{MOCHA}}, where relationships among tasks (clients) are separately modeled during the optimization of weights of tasks.
Our work is different from these methods in that we focus on the scenario of both few-shot and multi-task learning.

\section{Conclusion}

In this paper, we have presented \textsc{FewFedWeight}, enabling few-shot learning across massive NLP tasks with federated learning. 
In this new framework, the global model synthesizes pseudo samples for each client model, which are weighted by an energy-based     algorithm. 
Aggregation weights of client models are estimated according to their performance during training. 
Experiments on 118 different tasks demonstrate the effectiveness of the proposed \textsc{FewFedWeight}.

\clearpage
\bibliographystyle{named}
\bibliography{nips22}

\clearpage
\appendix

\section{Appendix}

\subsection{Task Type and Distribution} 
\label{appendix:split}

\begin{table}[t] 
\resizebox{0.95\textwidth}{!}{
\begin{tabular}{ll}
\hline
\textbf{Client} & \textbf{Tasks} \\
\hline
Client 1 & \makecell[l]{blimp-determiner-noun-agreement-with-adj-irregular-1, \\aeslc, glue-mrpc, math qa, quarel, kilt ay2,\\ 
tweet eval-stance atheism,  lama-squad, tab fact, aqua rat, \\tweet eval-emoji, glue-wnli, codah,  spider\\ 
tweet eval-offensive, wiki qa,  blimp-ellipsis n bar 1, \\openbookqa, sms spam, acronym identification,\\ 
ethos-national origin, definite pronoun resolution, \\hellaswag, superglue-wsc,  numer sense, \\ ade corpus v2-dosage, blimp-ellipsis n bar 2, \\squad-no context, e2e nlg cleaned }  \\
\hline
Client 2  & \makecell[l]{google wellformed query, xsum, wiqa, qa srl, \\tweet eval-stance abortion, ade corpus v2-effect, sick,\\ 
ethos-religion, commonsense qa, jeopardy, biomrc, \\superglue-multirc, ethos-race, eli5-askh, glue-qqp,\\ 
paws, ethos-directed vs generalized, glue-sst2, mocha, \\tweet eval-hate, glue-rte, hate speech offensive, \\
blimp-anaphor number agreement, lama-conceptnet, \\superglue-wic, boolq, kilt hotpotqa, aslg pc12, \\ quartz-no knowledge}  \\
\hline
Client 3 & \makecell[l]{tweet eval-stance climate, tweet eval-sentiment, qasc, \\medical questions pairs, break-QDMR-high-level, \\
imdb, glue-mnli, ethos-gender, trec-finegrained, \\crows pairs, adversarialqa, onestop english, duorc,\\
web questions, yelp review full, swag, proto qa, scitail, \\tweet eval-stance feminist, limit, common gen, \\
scicite, blimp-irregular past participle adjectives, \\social i qa, anli, kilt zsre, cosmos qa, superglue-record, \\
squad-with context}   \\
\hline
Client 4 & \makecell[l]{emotion, blimp-existential there quantifiers 1, sciq, \\race-middle, kilt wow, wino grande, rotten tomatoes, \\
superglue-cb, poem sentiment, ropes, piqa, quail, \\climate fever, lama-google re, search qa, wiki auto, \\
mc taco, blimp-wh questions object gap, hotpot qa, \\emo, kilt nq, kilt trex, quartz-with knowledge, aeslc,\\
dbpedia 14, yahoo answers topics, app reviews, \\superglue-copa, blimp-anaphor gender agreement, \\
hate speech18, gigaword}  \\
\hline
\end{tabular}
}
\caption{Tasks assigned to different clients.}
\label{tab:split}
\end{table}

The 118 tasks were randomly dispatched to 4 clients, which is shown in Table \ref{tab:split}. The 118 tasks can be categorized into 4 groups according to \citep{ye2021crossfit}: classification, QA, conditional generation and other. The categories of tasks are shown in Table \ref{tab:tasks_type}.

\begin{table}[t]
\resizebox{0.95\textwidth}{!}{
\begin{tabular}{ll}
\hline
\textbf{Type} & \textbf{Tasks} \\
\hline
Classification & \makecell[l]{anli, medical questions pairs, paws, glue-rte,\\
onestop english, poem sentiment, sick, glue-sst2,\\
scicite, rotten tomatoes, climate fever, glue-qqp,\\
scitail,  sms spam, dbpedia 14, emotion,  glue-wnli,\\
superglue-cb, emo, ethos-gender, imdb, glue-mrpc,\\
ethos-directed vs generalized, ethos-race, tab fact,\\
ethos-national origin, superglue-wic, superglue-wsc,\\
glue-mnli, google wellformed query, wiki qa,\\
hate speech18, hate speech offensive, ethos-religion,\\
trec-finegrained, tweet eval-emoji, wiki auto,\\
tweet eval-hate, tweet eval-offensive,\\
tweet eval-sentiment, tweet eval-stance abortion, \\
tweet eval-stance atheism, tweet eval-stance climate, \\
tweet eval-stance feminist, yahoo answers topics}  \\
\hline
QA  & \makecell[l]{lama-conceptne, adversarialqa, lama-google re, \\
lama-squad, math qa, mc taco, aqua rat, wino grande,\\
numer sense, biomrc, openbookqa, qasc, quail, \\
quarel, quartz-no knowledge, search qa, ropes,\\
quartz-with knowledge, race-middle, boolq, sciq, \\
codah, commonsense qa, jeopardy, social i qa, \\
squad-no context, duorc, squad-with context, \\
superglue-copa, eli5-askh, swag, superglue-multirc, \\
superglue-record, web questions, hellaswag,  wiqa, \\
kilt zsre, kilt nq, kilt ay2, kilt trex, cosmos qa \\
}  \\
\hline
Conditional Generation & \makecell[l]{aeslc, spider, gigaword, xsum, kilt wow}   \\
\hline
Other & \makecell[l]{acronym identification, ade corpus v2-dosage,\\
ade corpus v2-effect, limit, app reviews, crows pairs,\\
mocha, aslg pc12, blimp-anaphor gender agreement, \\
piqa, yelp review full, proto qa, common gen,\\ 
blimp-determiner noun agreement with adj irregular 1,\\
qa srl, blimp-ellipsis n bar 1, blimp-ellipsis n bar 2,\\
 blimp-existential there quantifiers 1, hotpot qa, \\
e2e nlg cleaned, blimp-anaphor number agreement, \\
blimp-irregular past participle adjectives,\\
definite pronoun resolution, break-QDMR-high-level, \\
blimp-wh questions object gap}  \\
\hline
\end{tabular}
}
\caption{The categories of 118 tasks.}
\label{tab:tasks_type}
\end{table}

\subsection{Meta Weighting} 
\label{appendix:metaweight}

\begin{small}
\begin{algorithm}[t]
	\caption{Meta Weighting}
	\label{alg:metaweight}
    \textbf{Input}: $\bm{X}_i, \bm{Y}_i, \hat{\bm{Y}}_i, \bm{F}_i, T_i, l$ \\
        
	\begin{algorithmic}[1] 
		\FOR{$t = 0,1,2,...,T_i-1$}
		\STATE Sample $\bm{B} \in (\bm{X}_i,\bm{Y}_i)$ and $\hat{\bm{B}} \in (\bm{X}_i,\hat{\bm{Y}}_i)$
		\STATE \textbf{First Step:}
        \STATE $\tilde{\bm{y}}_i \gets \text{Forward}(\bm{F}_i, \bm{x}_j)$, $\bm{x}_j \in \hat{\bm{B}}$
        \STATE $l_{\hat{\bm{B}}} \gets \sum_1^{b_i} l(\tilde{\bm{y}_j}, \hat{\bm{y}}_j)$
        \STATE $\bm{F}_i^{\prime} \gets \text{Optimize}(\bm{F}_i, l_{\hat{\bm{B}}})$
		\STATE \textbf{Second Step:}
        \STATE $\bar{\bm{y}}_j \gets \text{Forward}(\bm{W_i}^{\prime}, \bm{x}_j)$, $\bm{x}_j \in \bm{B}$
        \STATE $l_{\bm{B}}^{\prime} \gets \sum_1^{b_i} \bm{\epsilon}_j \cdot l(\bar{\bm{y}}_j, \bm{y}_j)$
        \STATE $\nabla{\bm{\epsilon}} \gets \frac{\partial l_{\bm{B}}^{\prime}}{\partial \bm{\epsilon}}$
        \STATE $\hat{\bm{w}} \gets \mathrm{max}(-\nabla{\bm{\epsilon}},0); \bm{w_j}=\frac{\hat{\bm{w}}_j}{\sum_1^{b_i} \hat{\bm{w}}_j}$
        \STATE \textbf{Third Step:}
        \STATE $\tilde{\bm{y}}_i \gets \text{Forward}(\bm{F}_i, \bm{x}_j)$, $\bm{x}_j \in \hat{\bm{B}}$
        \STATE $l_{\hat{\bm{B}}}^{\prime} \gets \sum_1^{b_i} \bm{w}_j\cdot l(\tilde{\bm{y}_j}, \hat{\bm{y}}_j)+l(\tilde{\bm{y}_j}, \bm{y}_j)$
        \STATE $\bm{F}_i^{t+1} \gets \text{Optimize}(\bm{F}_i, l_{\hat{\bm{B}}}^{\prime})$
        \STATE \textbf{Return} $\bm{F}_i^{t+1}$
		\ENDFOR
	\end{algorithmic}
\end{algorithm}
\end{small}

The meta weighting algorithm is shown in Algorithm \ref{alg:metaweight}. At the beginning, we randomly sample a mini-batch $\hat{\bm{B}} \in (\bm{X}_i, \hat{\bm{Y}}_i)$ and ${\bm{B}} \in (\bm{X}_i, \bm{Y}_i)$. In the first step, backpropagation is performed on $\hat{\bm{B}}$ and the client model $\bm{\bm{F}}_i$ is updated. We denote the original client model as $\bm{F}_i$ and the updated client model as $\bm{F}^{\prime}_i$. $\bm{F}^{\prime}_i$ will be evaluated on $\bm{B}$ to get the sample weight. 

In the second step, the loss function is changed. Assuming the original loss function is $l_{\bm{B}}=\sum_1^{b_i} l(\bar{\bm{y}}_j, \bm{y}_j)$, where $b_i$ is the batch size of $\bm{B}$ and $\hat{\bm{B}}$, $\bar{\bm{y}}_j$ is the prediction of $\bm{F}^{\prime}_i$ and $l$ is per-sample loss function. In this paper, we use CrossEntropy to compute $l$. The new loss function is computed as follows:
\begin{equation} \label{eq:loss1}
    \hat{l}_{\bm{B}}=\sum_{j=1}^{b_i} \bm{\epsilon}_j \cdot l(\bar{\bm{y}}_j, \bm{y}_j),
\end{equation}
where $\bm{\epsilon}_j$ is the perturbation to the per-sample loss. $\bm{\epsilon_j}$ is initialized to 1 for all samples. When $\bm{F}^{\prime}_i$ performs backpropagation on $\bm{B}$ with $\hat{l}_{\bm{B}}$, we get the grad to $\bm{\epsilon}$, $\nabla{\bm{\epsilon}}=\frac{\partial \hat{l}_{\bm{B}}}{\partial \bm{\epsilon}}$. The sample weight is computed as:
\begin{equation} \label{eq:sample_weight}
    \hat{\bm{w}} = \mathrm{max}(-\nabla{\bm{\epsilon}},0).
\end{equation}
To stabilize the training process, we normalize weights of samples from the same mini-batch as: $\bm{w}_j=\frac{\hat{\bm{w}}_j}{\sum_1^{b_i} \hat{\bm{w}}_j}$. 

In the third step, $\bm{F}_i$ is updated on $\bm{B}$ and $\hat{\bm{B}}$ with sample weights $\bm{w}$. The loss is computed as:
\begin{equation} \label{eq:loss2}
    l_{\hat{\bm{B}}}^{\prime}=\sum_{j=1}^{b_i} \bm{w}_j\cdot l(\tilde{\bm{y}_j}, \hat{\bm{y}}_j)+l(\tilde{\bm{y}_j}, \bm{y}_j),
\end{equation}
where $\tilde{\bm{y}}_j$ is the prediction of $\bm{F}_i$.

\end{document}